# A Review: PTSD in Pre-Existing Medical Condition on Social Media

Zaber Al Hassan Ayon[1], Nur Hafieza Ismail*[2], Nur Shazwani Kamarudin[3]

Faculty of Computing, University Malaysia Pahang Al-Sultan Abdullah, 26600 Pekan, Pahang, Malaysia[1, 2, 3]

*Abstract*—Post-Traumatic Stress Disorder (PTSD) is a multifaceted mental health condition, particularly challenging for individuals with pre-existing medical conditions. This review critically examines the intersection of PTSD and chronic illnesses as expressed on social media platforms. By systematically analyzing literature from 2008 to 2024, the study explores how PTSD manifests and is managed in individuals with chronic conditions such as cancer, heart disease, and autoimmune disorders, with a focus on online expressions on platforms like X (formally known as Twitter) and Facebook. Findings demonstrate that social media data offers valuable insights into the unique challenges faced by individuals with both PTSD and chronic illnesses. Specifically, natural language processing (NLP) and machine learning (ML) techniques can identify potential PTSD cases among these populations, achieving accuracy rates between 74% and 90%. Furthermore, the role of online support communities in shaping coping strategies and facilitating early interventions is highlighted. This review underscores the necessity of incorporating considerations of pre-existing medical conditions in PTSD research and treatment, emphasizing social media's potential as a monitoring and support tool for vulnerable groups. Future research directions and clinical implications are also discussed, with an emphasis on developing targeted interventions.

*Keywords*—PTSD; mental health; social media; natural language processing; health informatics

## I. INTRODUCTION

Mental health disorders, including conditions such as PTSD, depression, and anxiety, represent a substantial global public health challenge. The World Health Organization (WHO) estimates that one in four individuals will experience a mental disorder at some point in their lives, underscoring the widespread impact of these conditions on physical health and overall well-being [1]. Among these disorders, PTSD stands out due to its association with severe psychiatric comorbidities, including depression, anxiety, and an elevated risk of suicide, as well as its potential to exacerbate pre-existing physical health conditions [2].

PTSD is commonly understood to develop following exposure to traumatic events, such as warfare, natural disasters, or severe interpersonal violence, including sexual assault. Based on our analysis recent research has expanded this understanding to include PTSD triggered by serious medical challenges, such as chronic illnesses (e.g., cancer, heart disease, kidney failure, and lung disease) and complications during childbirth. Notably, studies suggest that up to 30% of individuals who experience life-threatening medical conditions may develop PTSD, highlighting the significant psychological toll of such experiences [3, 4, 5].

The onset of PTSD in medical contexts can be attributed to several factors, including the emotional shock of receiving a serious diagnosis, the physical and psychological demands of treatment, and the pervasive fear of death or disability. These experiences can lead to profound feelings of helplessness, loss of control, and vulnerability, all of which are key contributors to the development of PTSD [4, 6, 7]. Despite the growing recognition of PTSD in medical settings, there remains a critical need to understand how individuals with pre-existing medical conditions navigate and manage their PTSD symptoms.

In recent years, social media has emerged as a valuable tool for individuals with PTSD to seek support, share experiences, and access information. The impact of social media uses on PTSD symptoms, particularly in individuals with pre-existing medical conditions, is not yet fully understood [8]. The spontaneous and candid nature of social media posts provides a unique lens into individuals' thoughts, feelings, and behaviors, potentially revealing early signs of PTSD.

Advancements in artificial intelligence (AI), particularly in ML and NLP, have enabled the development of sophisticated algorithms capable of detecting patterns indicative of PTSD in social media data. These technologies hold promise for identifying PTSD cases with considerable accuracy, facilitating early intervention and personalized treatment approaches [9, 10]. During our analysis of the research works, the ethical implications and privacy concerns associated with using social media data for mental health research warrant careful consideration.

This article aims to provide a comprehensive overview of the intersection between PTSD and pre-existing medical conditions, focusing on the role of social media and ML in the identification and management of PTSD. By synthesizing current research, this review seeks to inform future research directions and clinical practices, ultimately improving outcomes for individuals grappling with both PTSD and chronic medical conditions.

To further underscore the significance of this study, the motivations include addressing the critical gap in understanding the interplay between PTSD and chronic medical conditions. This research aims to provide actionable insights that inform both clinical practices and the development of supportive technologies for affected individuals. This work highlights the transformative potential of social media analytics in mental health care, offering novel avenues for early intervention and monitoring.

*Corresponding Author

## II. MENTAL HEALTH DISORDERS

Mental disorders, as defined by the WHO, involve significant disturbances in cognition, emotional regulation, or behavior, leading to distress or impairment in daily functioning [1]. These conditions, which encompass a wide range of disorders, adversely affect cognitive abilities, emotions, and social interactions. Accurate and timely assessments are critical for proper diagnosis and intervention. Traditionally, mental health screening has relied on self-report questionnaires designed to identify specific symptoms or attitudes related to social interactions [11].

The recognition of mental illness through diverse data sources and algorithms has been a significant focus of recent research, with AI playing a pivotal role. ML models, trained on various datasets, including self-reported questionnaires and EEG (Electroencephalogram) data, have been employed to predict mental health disorders such as depression and PTSD. Early studies utilized datasets annotated by domain experts, while more recent research has explored the potential of social media data in detecting mental illness [12, 13, 14].

Technological advancements have revolutionized psychiatric research, enabling the rapid collection and analysis of vast datasets through mobile phones, sensors, and social media platforms. ML, particularly in the context of advanced statistical and probabilistic methods, has proven effective in detecting mental health conditions [15]. A study on mental health recognition the authors employed ML techniques to analyze bipolar disorder using the Mood Disorder Questionnaire, while another study achieved 98.6% accuracy in detecting stress levels from bio-signals using supervised ML [11, 16]. Similarly, another study demonstrated a model for assessing psychiatric distress using EEG signals, employing algorithms such as Support Vector Machine (SVM), Logistic Regression (LR), and Naive Bayes (NB), with notable accuracy rates in stress detection [17].

In addition to ML, deep learning (DL) techniques have emerged as powerful tools in predicting depression risk. Context-DNN (Context Deep Neural Network) models, for instance, have been used to estimate the likelihood of depression [18]. Social media platforms, as a repository of user-generated content, have become valuable resources for observing mental states and psychiatric disorders. Studies have utilized platforms such as Reddit to analyze mental illness-related content through DL approaches, diagnosing chronic mental disorders like panic, bipolar disorder, and ADHD (attention deficit hyperactivity disorder) [19].

Other research has focused on detecting various mental illnesses, including schizophrenia, bipolar disorder, and PTSD, using data from platforms such as X, Facebook, and Weibo [18, 20]. Recent advancements in Large Language Models (LLMs) [86] have shown promise in recognizing mental health disorders from social media interactions, opening new possibilities for early identification and intervention [21]. LLMs are being explored as a tool to identify potential mental health concerns by analyzing text and speech patterns [22]. LLMs can be trained on large datasets of social media posts to identify patterns of language use associated with specific mental health conditions [23]. Regardless of the recent advancement of LLM in mental health identification, research and development are crucial to ensure accuracy and reliability before these tools can be integrated into clinical practice.

The expansion of social media has created unparalleled opportunities for mental health research, offering real-time insights into individuals' mental states. Advances in NLP and ML have enabled the identification of patterns in social media data that are indicative of mental health conditions, such as depression, anxiety, and PTSD. Studies have demonstrated the potential of these technologies to reveal linguistic markers and behavioral patterns associated with mental health issues, with predictive accuracy [6, 24]. Despite these advancements, the use of social media data for mental health research raises ethical concerns, including privacy issues, the risk of misdiagnosis, and potential stigmatization. The representativeness of social media data and the accuracy of self-reported mental health statuses remain areas of debate. Nonetheless, the integration of social media analysis into mental health research offers significant potential for early detection and intervention, potentially transforming the field of mental health care in the digital age.

Recent research has underscored the efficacy of combining social media data with LLMs for the diagnosis of mental health disorders, particularly in the early stages. The reluctance of many individuals to undergo mental health evaluations, coupled with the growing use of social media to share personal experiences, presents a unique opportunity for early diagnosis through social media analysis. The spontaneous and candid nature of social media expressions, when analyzed by sophisticated language models, can reveal subtle indicators of mental health issues that might otherwise go undetected. This approach holds significant promise for revolutionizing early intervention strategies in mental health care, potentially leading to improved outcomes through timely diagnosis and treatment initiation.

## III. PTSD IN PRE-MEDICAL CONDITIONS

PTSD is a complex psychiatric condition that arises from exposure to traumatic events or prolonged distressing circumstances [25]. Traditionally associated with combat veterans and survivors of violence or natural disasters, PTSD also manifests in patients with pre-existing medical conditions, posing significant challenges for both patient care and treatment outcomes. The intersection of PTSD and chronic illness is an increasingly important area of study, with far-reaching implications for healthcare practices and economics.

The prevalence of PTSD is substantial, with approximately 3.5% of U.S. adults affected annually, and lifetime prevalence reaching 8% among adolescents aged 13-18 [26]. Globally, data from the World Mental Health Survey Consortium reveal that over 70% of the population has experienced at least one traumatic event, highlighting the widespread potential for PTSD development [27]. In the medical context, patients diagnosed with conditions such as cancer, heart disease, and chronic respiratory illnesses are at heightened risk for PTSD, with the psychological burden of these conditions creating fertile ground for trauma-related stress responses [28, 29]. This bidirectional relationship, where PTSD exacerbates physical symptoms and impairs treatment outcomes, underscores the importance of recognizing and addressing PTSD within medical populations [30].

The Coronavirus disease (COVID-19) pandemic has further emphasized the link between PTSD and medical conditions, with survivors of severe infections showing PTSD symptoms, particularly among those with risk factors like a history of psychiatric disorders or experiences of delirium during illness [28]. The pandemic itself is a traumatic event capable of inducing PTSD, as evidenced by studies showing high prevalence rates due to factors such as lockdowns, economic instability, and social isolation [21, 31].

Another critical area of concern is childbirth-related PTSD (CB-PTSD). Approximately 6% of women experience CB-PTSD, affecting about eight million women globally in 2022 [32, 33]. High-risk factors include medically complicated deliveries, obstetrical complications, and maternal near-miss incidents [34, 35]. Racial and ethnic disparities exacerbate these risks, with Black and Latin women nearly three times more likely to exhibit acute stress responses to childbirth [32]. Overall, about 20% of high-risk individuals are likely to develop CB-PTSD [32].

The Diagnostic and Statistical Manual of Mental Disorders, 5th edition (DSM-5), identifies core PTSD symptoms such as intrusive memories, avoidance, negative alterations in cognition and mood, and heightened arousal [36]. In medical contexts, these symptoms may manifest as flashbacks to painful procedures, avoidance of necessary treatments, and negative health beliefs, complicating medical management [37, 38]. Early detection and trauma-informed interventions are thus crucial in these settings to prevent misdiagnosis and ensure comprehensive care [30].

While effective treatments for PTSD, including psychotherapy and pharmacotherapy, are available, co-occurring conditions such as depression and substance use disorders complicate recovery. Addressing both psychological and physiological aspects of trauma through integrated care approaches is essential for improving patient outcomes [39].

The interplay between PTSD and pre-existing medical conditions warrants ongoing research and clinical focus. Future studies should aim to develop tailored interventions for specific medical populations, explore the neurobiological mechanisms of medical trauma-induced PTSD, and assess the long-term effectiveness of integrated care models. Advancing our understanding in this area will contribute to more compassionate and effective care for those grappling with the dual challenges of medical conditions and PTSD [40].

## IV. MENTAL HEALTH ASSESSMENT FOR DIAGNOSIS

Mental health assessment is a fundamental component of diagnosing and managing psychological and psychiatric disorders, requiring a multifaceted approach that integrates neuroimaging, physiological and laboratory analyses, clinical interviews, psychometric tools, and increasingly, social media data analysis [41]. This comprehensive methodology provides a nuanced understanding of the interplay between neurobiological, physiological, psychosocial, and digital behavioral factors underlying mental health conditions.

Neuroimaging techniques, particularly functional magnetic resonance imaging (fMRI) and positron emission tomography (PET), have become essential in identifying the neural substrates associated with various psychiatric disorders. For example, depressive disorders often involve reduced activity in the prefrontal cortex and increased amygdala activation, while schizophrenia is linked to anomalies in frontal and temporal lobe function [42]. These neurobiological insights are crucial for developing targeted and personalized therapeutic interventions [40].

Physiological and laboratory analyses complement neuroimaging by identifying underlying medical conditions that may contribute to psychiatric symptoms, such as thyroid dysfunction or nutritional deficiencies [43]. Comprehensive medical evaluations, including hematological profiles, endocrine function tests, and genetic analyses, are instrumental in guiding effective treatment strategies tailored to individual needs [43, 44].

Clinical interviews and standardized psychometric instruments remain central to mental health assessment. These methods allow clinicians to systematically gather detailed information on an individual's symptoms, personal and familial history, and functional impairments across various domains of life. The mental status examination, a critical component of the clinical interview, evaluates cognitive and behavioral domains such as affect, thought processes, and memory functions [45], offering crucial insights into the individual's psychological state and potential neuropsychiatric impairments.

Recently, the analysis of social media data has emerged as a promising addition to mental health assessment. The widespread use of social media platforms provides researchers and clinicians with real-time, naturalistic data on individuals' thoughts, emotions, and behaviors. This digital footprint can offer valuable insights into mental health trajectories and conditions.

NLP and ML algorithms have been developed to analyze social media content for markers of mental health conditions. For instance, changes in social media activity, linguistic style, and emotional expression have been used to predict the onset of depression with considerable accuracy [6]. Similarly, researchers have demonstrated the potential to diagnose PTSD from X data [24].

The integration of social media data analysis into mental health assessment offers several advantages:

- Early detection: Social media analysis can identify subtle changes in behavior or language indicative of emerging mental health issues before clinical symptoms appear.

- Continuous monitoring: Unlike traditional assessments, social media provides a continuous stream of data,

allowing for dynamic and responsive monitoring of mental health.

- Ecological validity: Social media reflects individuals' behaviors in natural environments, offering potentially more valid insights than clinical settings.

- Reach and accessibility: Social media analysis can extend mental health screening to populations that may not access traditional mental health services.

Nevertheless, the use of social media data in mental health assessment also presents significant ethical and practical challenges, including concerns about privacy, consent, data security, and the potential for misinterpretation. Additionally, biases in social media use and algorithmic analysis could lead to disparities in assessment and diagnosis [46, 47].

The integration of social media data with traditional assessment methods requires careful validation and robust ethical guidelines. Ongoing research is focused on refining algorithms, establishing normative data, and developing best practices for the responsible use of social media data in mental health assessment [46].

## V. PTSD Diagnosis on Social Media

The advent of social media has markedly transformed how individuals express and manage mental health conditions, including PTSD. While traditional diagnostic methods remain essential, social media platforms offer novel avenues for researchers and clinicians to explore the lived experiences of those with PTSD. This is particularly crucial given the high prevalence of PTSD among veterans, with 15-20% affected, highlighting the need for innovative approaches to address this critical issue [5].

Historically, PTSD identification in populations such as cancer survivors has relied on questionnaires, but these are time-consuming and impractical for large-scale studies. Methods like fMRI, though informative, are cost-prohibitive for widespread use. An alternative approach involves analyzing public social media posts, which offer quick, accessible data from a broad audience. A pioneer study on PTSD in cancer patients reports that approximately 60% of adults use online resources for health information, and social media platforms allow individuals to discuss health concerns more openly than in face-to-face interactions [20]. Previous studies have utilized platforms like Reddit to detect early indicators of mental health disorders, while others have analyzed Twitter data to understand language patterns among PTSD patients with a history of cancer [7, 46, 47].

The challenge of accurately diagnosing PTSD, particularly in cancer survivors, is compounded by the lack of measurable data, a gap that motivates the collection of social media data for analysis. Various statistical methods, including correlations, chi-square tests, and regression analyses, have been employed to evaluate mental health datasets, revealing variables closely associated with mental health diagnoses [48]. ML algorithms, both supervised and unsupervised, have proven effective in tracking mental illness symptoms with high diagnostic accuracy. For instance, linear discriminant analysis has been used to explore social media content, identifying topics relevant to mental health [7, 24].

Recent studies highlight the potential of social media analysis to identify symptoms like anger associated with PTSD, particularly within the veteran community. This method shows promise as a preventative measure, enabling early detection and timely intervention. The complex nature of military service, including exposure to combat and the difficulty in distinguishing combatants from non-combatants, contributes to the high prevalence of PTSD among veterans, underscoring the urgency of exploring new strategies like social media analysis for supporting military personnel's mental health [6].

Advanced ML techniques, such as recurrent neural networks (RNN), deep neural networks (DNN), and convolutional neural networks (CNN), are increasingly employed for text processing in mental health research. Sentiment analysis on X data, for example, has been suggested for understanding the context of communication and dialogue related to mental health [49].

In a novel approach, recent research has explored the detection of CB-PTSD using LLMs like Bio_ClinicalBERT and BioGPT, which have outperformed ChatGPT in clinical tasks. This study reveals that approximately 6% of the global childbearing population, or over eight million women annually, develop CB-PTSD, significantly impacting mothers and their children. The generative AI model such as Open AI text embedding (ADA), by analyzing maternal narratives, achieved an F1 score of 0.81 in identifying PTSD, indicating its potential to generalize to other mental health disorders [27].

As mental health research continues to evolve, the integration of social media data analysis holds significant potential for enhancing our understanding of PTSD. Leveraging user-generated content on these platforms allows researchers to gain deeper insights into PTSD symptoms, triggers, and coping mechanisms. This real-time, granular data can complement traditional survey methods, facilitating the early identification of mental health issues and the development of personalized interventions. Additionally social media analysis can inform public health initiatives, guiding resource allocation and outreach programs to address the mental health needs of vulnerable populations, particularly veterans and active-duty military personnel.

## VI. Text Classification in PTSD Identification

Accurately classifying and diagnosing PTSD in medical settings remains a formidable challenge, with significant implications for patient outcomes and healthcare resource allocation. PTSD, a prevalent psychiatric disorder, especially among military personnel and veterans, manifests in severe symptoms that drastically impair quality of life [4].

This review addresses the complexities of PTSD classification and diagnosis by exploring key processes, starting with the identification of robust data sources, such as clinical records, screening questionnaires, and social media content. These diverse data sources provide a comprehensive foundation for analyzing PTSD symptoms. Following data collection, precise annotation is crucial, as it involves accurately labelling symptoms essential for training effective models. Feature selection further refines the dataset, ensuring that the most relevant PTSD indicators are highlighted, thus enhancing model accuracy.

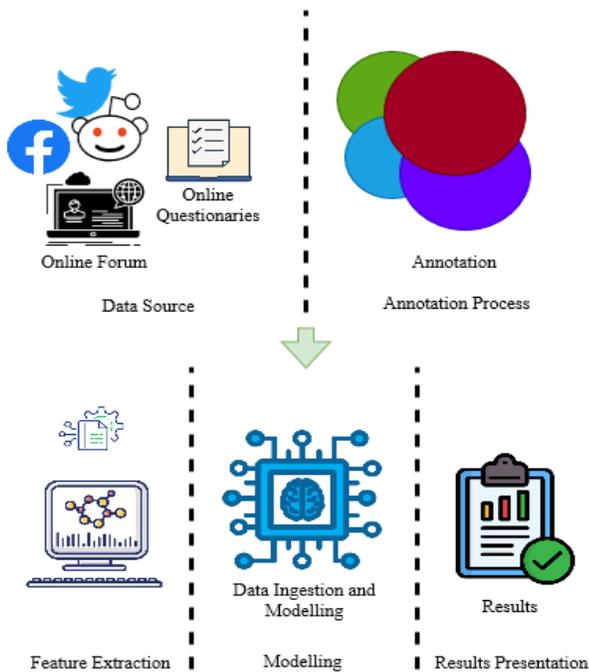

Fig. 1. Generic representation of PTSD identification from social media data.

Fig. 1 depicts a general approach followed by the reviewed studied related to identification of mental health disorder from social media data. As data source most of the studies. All most every study mentioned about human annotation via domain experts. Finally descriptive or inferential model training and feature extraction processes were followed to get class probability distribution.

Advanced modelling techniques, particularly ML algorithms, are then applied to predict PTSD presence. These models undergo rigorous testing to ensure robustness, with results critically analyzed to evaluate their effectiveness in medical settings. This process not only offers insights into the practical application of these models but also identifies potential areas for further refinement and research, ultimately contributing to more accurate PTSD diagnoses and improved patient care.

*A. Data Sources*

In recent years, researchers have utilized a range of data collection methods to explore the intersection of PTSD, pre-existing medical conditions, and social media engagement. This review synthesizes the key data sources employed in this burgeoning field, critically examining their strengths, limitations, and overall contributions to advancing our understanding of this complex and multifaceted issue. By integrating findings from diverse methodologies, this review seeks to provide a comprehensive overview of how these data sources have informed current research and to identify potential avenues for future inquiry.

*1) Questionnaires:* Several studies have employed online self-administered screening questionnaires to evaluate the prevalence of PTSD symptoms among individuals with pre-existing medical conditions who are active on social media platforms [24, 50]. These questionnaires are designed to collect comprehensive data, including demographic information, medical history to identify pre-existing conditions, and responses to validated PTSD screening tools such as the PTSD Checklist for Diagnostic and Statistical Manual of Mental Disorders, Fifth Edition (DSM-5) also known as PTSD Checklist Fifth Edition (PCL-5) Additionally, they include questions regarding social media usage and online behavior, enabling researchers to examine the relationship between PTSD symptoms and social media engagement.

The primary advantage of this approach is its capacity to generate quantitative data on the prevalence of PTSD symptoms and to explore the potential correlation between these symptoms and social media use. Even though it is important to acknowledge the methodological limitations inherent in this approach, particularly the risk of self-selection bias and the potential for underreporting or overreporting of symptoms within online environments. These factors must be carefully considered when interpreting the results to ensure the validity and reliability of the findings.

*2) Online discussion sites:* Numerous studies have investigated online discussion sites, including mental health forums and support groups, to gain insights into the experiences of individuals dealing with PTSD alongside pre-existing medical conditions [51, 52]. Key data sources for these studies include Reddit communities (subreddits) focused on PTSD and chronic illnesses, established mental health support forums, and condition-specific forums where mental health discussions are prevalent. These platforms provide rich qualitative data that allow for a nuanced analysis of personal narratives and coping mechanisms. The anonymity inherent to these platforms poses challenges in verifying the authenticity of the reported experiences.

*3) Social media platforms:* Individuals with chronic conditions are more likely to engage with health-related content on social media, while those with mental health conditions tend to engage less frequently [53]. An expanding body of research has focused on analyzing the social media activity of individuals who publicly discuss their mental health issues, including PTSD, in conjunction with pre-existing medical conditions [6, 54]. These studies typically investigate X posts using relevant hashtags, public Facebook groups dedicated to PTSD and chronic illness support, and Instagram posts with captions referencing both PTSD and other medical conditions.

This methodological approach provides valuable insights into how individuals publicly navigate their experiences with PTSD and pre-existing medical conditions on social media platforms. Researchers must also carefully consider ethical concerns related to privacy and consent when utilizing publicly available data, as these factors are critical to maintaining the integrity and ethical standards of the research [55].

The triangulation of diverse data sources has emerged as a best practice in this field, enabling researchers to obtain a comprehensive understanding of how PTSD manifests and is discussed within the context of pre-existing medical conditions on social media platforms [56, 57]. This multi-faceted approach facilitates both the quantitative assessment of symptom prevalence and the qualitative analysis of lived experiences and online behaviors.

Despite the benefits of these methods, researchers face several challenges when utilizing these data sources. First, representativeness is a significant concern; social media users may not accurately reflect the broader population of individuals with PTSD and pre-existing medical conditions [58]. For instance, Biernesser et al. found that social media data often over represent younger, more technologically savvy demographics, potentially leading to skewed results [58]. Second, the quality of data is a critical issue, as the reliability and validity of self-reported information on social media platforms can be questionable. This necessitates rigorous verification and cross-referencing of data, as highlighted by a study conducted by Torous et al., which emphasized the need for robust data cleaning and validation processes in mental health research using social media data [59].

Ethical considerations also play a crucial role, particularly in balancing the need for research with privacy concerns and informed consent. Golder et al. proposed a framework for ethical social media research in health contexts, emphasizing the importance of protecting user privacy and obtaining appropriate consent [60]. Additionally, technological barriers must be considered; the rapid evolution of social media platforms and their algorithms can impact data collection and analysis methods, requiring researchers to continuously adapt their approaches. Staying informed about platform changes and their implications for data accessibility and analysis is essential [61].

Future research in this domain would benefit from the development of standardized protocols for social media data collection and analysis, alongside the exploration of innovative approaches to integrating online and offline data sources. Such integration could yield a more comprehensive understanding of PTSD in the context of pre-existing medical conditions. Conway and O'Connor on their study suggest that combining traditional clinical data with social media insights could enhance the robustness and generalizability of findings [62]. By critically examining these diverse data sources and their applications, researchers can refine their methodologies and contribute meaningfully to the expanding body of literature at the intersection of mental health, chronic illness, and digital spaces.

*B. Data Annotation*

Data annotation is pivotal in the analysis of social media content for PTSD and other mental health conditions, playing a crucial role in ensuring the validity and reliability of research findings. This process involves several key methods and tools, each contributing uniquely to the field. A fundamental aspect of data annotation is the identification and labelling of specific PTSD symptoms in social media posts. Researchers utilize a variety of techniques for this task, ranging from NLP and ML algorithms to manual coding by trained clinicians and keyword-based approaches [6, 24, 63]. These methods are designed to detect symptoms such as re-experiencing, avoidance, negative alterations in cognition and mood, and heightened arousal, with each approach balancing considerations of efficiency, accuracy, and scalability.

Another critical component of this process is establishing the ground truth for data collection, which involves validating annotated data against reliable sources. This validation can be achieved through clinical verification via interviews, self-reported PTSD symptom measures, or comparisons with clinician-diagnosed cases [49, 64]. Human annotators continue to play an essential role in this process, providing a nuanced understanding and insights that may be difficult to capture through automated methods alone. Expert annotators, such as mental health professionals, offer high-quality annotations based on clinical expertise, while trained lay annotators and crowdsourcing platforms provide scalable alternatives, albeit with the necessity of rigorous quality control [9, 65].

To deepen the analysis, researchers frequently incorporate additional tools and measures. Depression survey scores, such as the Patient Health Questionnaire-9 (PHQ-9), are often used alongside PTSD assessments to evaluate symptom severity and understand comorbidities [6, 66]. The Linguistic Inquiry and Word Count (LIWC) tool has also gained prominence for its ability to analyze linguistic patterns in social media posts, offering valuable insights into the emotional and cognitive processes associated with PTSD [49, 63].

Recently, the advent of LLMs as annotators has introduced new possibilities in data annotation for PTSD research. LLMs such as GPT-3 and Bidirectional Encoder Representations from Transformers (BERT) show considerable promise in automating symptom detection and efficiently processing vast amounts of data [67, 68, 69]. However, their use raises important considerations. Ethical concerns related to privacy and consent, the inherent lack of clinical expertise in these models, limitations in understanding broader contextual nuances, and issues of reliability and transparency present significant challenges. While LLMs offer potential benefits in terms of scalability and efficiency, their application in mental health contexts requires careful validation against established clinical standards and ongoing scrutiny to ensure ethical and accurate annotations.

In conclusion, the process of data annotation for PTSD research on social media is continuously evolving, with each method presenting distinct strengths and limitations. A balanced approach that integrates human expertise with technological advancements appears most promising. As the field advances, researchers must skillfully navigate the complexities of various annotation methods, striving for a harmonious integration that ensures both efficiency and clinical relevance in the pursuit of understanding PTSD through social media data.

## C. Feature Selection

In the context of analyzing PTSD in individuals with pre-existing medical conditions on social media, feature selection is critical for identifying the most relevant characteristics that enhance the effectiveness of predictive modelling. Researchers utilize a range of techniques to distil meaningful features from the extensive data available on social media platforms. Commonly extracted textual features include word frequency, n-grams, and sentiment scores derived from posts and comments, which provide insights into the language patterns associated with PTSD [24]. Temporal patterns of social media activity, such as posting frequency and the time-of-day posts are made, have also proven significant in detecting PTSD, offering a behavioral dimension to the analysis [6].

In addition to these features, researchers frequently incorporate user profile information and engagement metrics, which may serve as potential indicators of PTSD symptoms. Advanced approaches further enhance the analysis by leveraging NLP techniques to capture semantic and contextual features, such as topic modelling and word embeddings, allowing for a deeper understanding of the content and its relevance to PTSD [69].

The primary challenge in feature selection lies in balancing the richness of these features with the risk of overfitting, particularly given the complex and multifaceted nature of PTSD and its interactions with pre-existing medical conditions. To address this, dimensionality reduction techniques like Principal Component Analysis (PCA) and t-SNE are often employed. These methods help manage high-dimensional feature spaces by condensing the data while preserving the most critical information [70].

Recent studies have explored the use of automated feature selection methods, including wrapper and embedded techniques, which optimize the feature set for detecting PTSD in the context of comorbid conditions. These methods offer a systematic approach to refining feature selection, thereby improving model accuracy and robustness [71].

By carefully selecting and refining features, researchers can build more effective models that not only predict PTSD but also account for the complexity of its interactions with other medical conditions, ultimately contributing to more accurate and meaningful insights in mental health research.

## D. Modeling

The modelling approaches employed in studying PTSD among individuals with pre-existing medical conditions on social media encompass a broad spectrum of techniques, ranging from traditional statistical methods to cutting-edge ML and DL algorithms.

Statistical methods have long served as the cornerstone of research in this field, providing foundational tools for analysis. LR remains a widely favored approach due to its interpretability and its capability to quantify the relationship between various features and the likelihood of PTSD [24]. Additionally, survival analysis techniques, particularly Cox proportional hazards models, are frequently used to explore the temporal dynamics of PTSD development in individuals with pre-existing conditions, identifying key risk factors over time [16]. Multilevel modelling has gained prominence for its ability to account for the nested structure of social media data—such as posts within users and users within platforms—thereby accommodating the hierarchical nature of the data and enhancing the accuracy of the analysis.

ML methods have also been pivotal in advancing PTSD detection and risk assessment. SVMs, known for their effectiveness in handling high-dimensional feature spaces, have been successfully applied to classify social media users with PTSD [13, 14]. Random Forests and Gradient Boosting Machines offer robust performance by managing complex interactions between features, making them particularly valuable in assessing the interplay between PTSD and co-existing medical conditions [41] [85]. Ensemble methods, which combine the strengths of multiple algorithms, have demonstrated superior accuracy and generalizability in PTSD prediction models [12].

The rise of DL has introduced powerful tools for analyzing the unstructured and complex nature of social media data in PTSD research. CNNs have been utilized to detect PTSD-related language by capturing local patterns in text data, demonstrating their efficacy in this domain [17]. RNNs, especially Long Short-Term Memory (LSTM) networks, excel in modeling temporal dependencies in social media activity, making them well-suited for identifying patterns associated with PTSD [23] Transformer-based models, such as BERT and its variants, have emerged as particularly effective in understanding the nuanced contexts in which PTSD expressions occur in social media posts. These models, when fine-tuned on domain-specific data, can capture subtle linguistic cues indicative of PTSD [19]. Additionally, the use of graph neural networks is gaining traction, as they can model the complex relationships between users, posts, and symptoms in social networks, offering a novel perspective on understanding PTSD within the dynamics of online social interactions [34].

Recently, LLMs have revolutionized PTSD research on social media by providing advanced capabilities for feature extraction and modelling. Pre-trained on vast text corpora, these models bring a sophisticated understanding of language and context, which is essential for identifying PTSD. Researchers have explored several applications of LLMs in this area. For instance, fine-tuning approaches adapt pre-trained models like BERT to the specific task of detecting PTSD in social media texts, capturing subtle linguistic and contextual nuances associated with PTSD symptoms [22]. Another approach leverages LLMs as feature extractors, utilizing the rich representations learned by these models as inputs for other ML algorithms [10]. Recent studies, such as those by Yang et al. have developed frameworks using ChatGPT and LLaMA-2 to automate PTSD assessments from clinical interviews, highlighting the potential of LLMs to enhance diagnostic accuracy [23]. Additionally, a group of researchers introduced innovative text augmentation methods using LLMs to address data imbalance issues in PTSD diagnosis, further demonstrating the versatility and impact of these models [72].

In summary, the evolution of modeling approaches in PTSD research on social media reflects a dynamic interplay between traditional and modern techniques. By integrating statistical methods, ML, DL, and LLMs, researchers can achieve a more nuanced and comprehensive understanding of PTSD in individuals with pre-existing medical conditions, ultimately advancing the field's ability to accurately detect and model this complex disorder. Fig. 2 depicts the general approach followed by many studies to implement AI modeling with NLP.

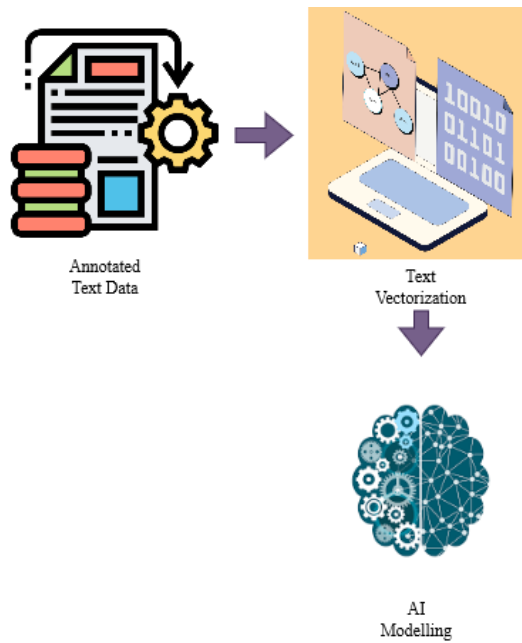

Fig. 2. Generic artificial intelligence predictor modelling.

*E. Results*

The study of PTSD in individuals with pre-existing medical conditions using social media data has yielded significant insights through various modeling approaches. Statistical methods, including survival analyses, have identified critical time windows for intervention following diagnoses of conditions like chronic pain and autoimmune disorders, emphasizing the need for timely support [73, 74, 75]. Traditional ML models, such as SVMs and Random Forests, have demonstrated strong performance, with accuracy rates between 75-85% and AUC scores exceeding 0.90 in some cases [24, 76, 77]. The variability in performance across different conditions indicates that these models require further tuning for condition-specific applications. DL techniques, such as CNNs, LSTMs, and Transformer-based models like BERT, have advanced PTSD detection by capturing nuanced language patterns in social media posts, achieving precision rates up to 88% and F1 scores between 0.85-0.90 [9, 37]. The "black box" nature of these models presents challenges for clinical application due to difficulties in interpretability [78].

Sentiment analysis tools, such as the LIWC tool and the VADER (Valence Aware Dictionary for Sentiment Reasoning) tool, have shown promise in identifying mental health conditions. LIWC has revealed significant differences in the linguistic style of individuals experiencing emotional distress compared to those who are not, suggesting its utility in mental health detection [19]. VADER, known for its computational efficiency and extensive word corpus, has been used effectively to predict depression risk based on sequential social media messages, demonstrating the evolving nature of sentiment analysis in this field [79]. Feature engineering methods, including bag-of-words and N-grams, combined with these sentiment analysis tools, have achieved accuracy rates ranging from 74% to 82%, suggesting that text-based screening tools hold substantial potential for identifying individuals at risk of PTSD or related mental health conditions [79, 80].

LLMs, such as GPT-3 and text-embedding-ada-002, have shown impressive accuracy in detecting PTSD, with F1 scores and accuracy rates up to 0.82 and 83%, respectively [81, 82]. Despite the progress, challenges remain, particularly with the "black box" nature of advanced models, computational demands, and privacy concerns [83, 84]. Traditional ML and sentiment analysis tools provide a good balance of accuracy and interpretability but often lack the depth to capture complex PTSD manifestations. DL models and LLMs offer superior performance in detecting nuanced expressions of trauma but suffer from interpretability and ethical concerns, which limit their practical use in clinical settings. The integration of these methods with a focus on improving interpretability and ethical use is crucial for advancing PTSD detection and intervention strategies.

## VII. CONCLUSION AND FUTURE WORK

This review has examined the intersection of PTSD, pre-existing medical conditions, and social media behavior, highlighting the significant advancements in the field from basic keyword analysis to the application of sophisticated ML and LLMs. These developments demonstrate substantial potential for the early detection and intervention of PTSD among individuals with chronic health conditions.

Despite these advancements, several critical knowledge gaps persist. First, there is currently no consensus on best practices for data collection, annotation, and analysis in social media-based PTSD research, particularly in the context of comorbid conditions. Additionally, most of the existing research relies on cross-sectional data, leaving the temporal dynamics of PTSD development in individuals with pre-existing conditions inadequately understood. Furthermore, a significant disconnect exists between insights derived from social media and their application in clinical practice, limiting the practical utility of research findings. The specific interactions between certain medical conditions and PTSD manifestations on social media are also not sufficiently studied. Comprehensive ethical guidelines for the use of AI in mental health detection on social media are lacking, raising concerns about privacy, consent, and the responsible use of predictive models.

To address these gaps, future research should prioritize the following areas. First, establishing clear guidelines for data collection, annotation, and analysis is essential to enhance the reproducibility and comparability of studies across the field. Collecting users single post and training an inferential model for prediction might not give the model full understanding of sarcasm, jokes, humor due incomplete reasoning over the dataset. Second, designing long-term studies that track PTSD development in individuals with pre-existing conditions, using continuous social media data, will provide deeper insights into the temporal dynamics of PTSD. Third, combining social media data with electronic health records and data from wearable devices can offer a more comprehensive understanding of PTSD in the context of comorbid conditions.

Fourth, creating adaptive models that account for individual differences and specific interactions between pre-existing conditions and PTSD is crucial for enhancing the accuracy and relevance of predictive tools. Fifth, developing methods to make advanced models, particularly LLMs, more interpretable for clinical application is vital for their effective integration into healthcare settings. Sixth, expanding research to include diverse cultural contexts will help in understanding varying PTSD manifestations and social media usage patterns across different populations. Lastly, establishing robust ethical frameworks for the use of AI in mental health contexts is necessary to address issues of consent, privacy, and the responsible deployment of predictive models.

In conclusion, while social media analysis offers unprecedented opportunities for understanding PTSD in the context of pre-existing medical conditions, realizing its full potential requires addressing these significant knowledge gaps. Future research should focus on standardization, longitudinal studies, and the development of ethical, interpretable AI models that can be effectively integrated into clinical practice. Interdisciplinary collaboration will be crucial in translating these research insights into tangible improvements in mental health care for individuals living with both PTSD and chronic medical conditions.

## ACKNOWLEDGMENT

This research was fully funded by UMPSA Research Grant Scheme under grant RDU220350, UMPSA.